%% file: Formatting-Instructions-LaTeX-2026.tex
\documentclass[letterpaper]{article} 
\usepackage{aaai2026}  
\usepackage{times}  
\usepackage{helvet}  
\usepackage{courier}  
\usepackage[hyphens]{url}  
\usepackage{graphicx} 
\usepackage{natbib}  
\usepackage{caption} 
\usepackage{algorithm}
\usepackage{multirow}
\usepackage{longtable}
\usepackage{xltabular}

\usepackage{booktabs}
\usepackage{enumitem}
\usepackage{amsmath}
\usepackage{algpseudocode}
\usepackage{amssymb}
\usepackage[utf8]{inputenc}
\usepackage[most]{tcolorbox}

\newcommand{\our}{CDCR‑SFT}
\usepackage{newfloat}
\usepackage{listings}
\DeclareCaptionStyle{ruled}{labelfont=normalfont,labelsep=colon,strut=off} 
\floatstyle{ruled}
\newfloat{listing}{tb}{lst}{}
\floatname{listing}{Listing}
\title{Mitigating Hallucinations in Large Language Models via Causal Reasoning}

\author{
    Yuangang Li\textsuperscript{\rm 1,\thanks{These authors contributed equally.}}, 
    Yiqing Shen\textsuperscript{\rm 2,\addtocounter{footnote}{-1}}\footnotemark, 
    Yi Nian\textsuperscript{\rm 1},   
    Jiechao Gao\textsuperscript{\rm 3}, 
    Ziyi Wang\textsuperscript{\rm 4}, 
    Chenxiao Yu\textsuperscript{\rm 1},\\ 
    Shawn Li\textsuperscript{\rm 1},
    Jie Wang\textsuperscript{\rm 3}, 
    Xiyang Hu\textsuperscript{\rm 5,\thanks{Corresponding authors.}}, 
    Yue Zhao\textsuperscript{\rm 1,\addtocounter{footnote}{-1}}\footnotemark\\
}
\affiliations{
    \textsuperscript{\rm 1}University of Southern California, \textsuperscript{\rm 2}Johns Hopkins University, \textsuperscript{\rm 3}Stanford University\\
    \textsuperscript{\rm 4}University of Maryland, College Park, \textsuperscript{\rm 5}Arizona State University\\
    
\{yuangang,yinian,cyu96374,li.li02,yue.z\}@usc.edu, yshen92@jhu.edu,\\
\{jiechao,jiewang\}@stanford.edu, zoewang@umd.edu, xiyanghu@asu.edu
}

\usepackage{bibentry}

\begin{document}

\maketitle

\begin{abstract}
Large language models (LLMs) exhibit logically inconsistent hallucinations that appear coherent yet violate reasoning principles, with recent research suggesting an inverse relationship between causal reasoning capabilities and such hallucinations. 
However, existing reasoning approaches in LLMs, such as Chain-of-Thought (CoT) and its graph-based variants, operate at the linguistic token level rather than modeling the underlying causal relationships between variables, lacking the ability to represent conditional independencies or satisfy causal identification assumptions.
To bridge this gap, we introduce Causal-DAG construction and reasoning (\our{}), a supervised fine-tuning framework that trains LLMs to explicitly construct variable-level directed acyclic graph (DAG) and then perform reasoning over it.
Moreover, we present a dataset comprising 25,368 samples (CausalDR), where each sample includes an input question, explicit causal DAG, graph-based reasoning trace, and validated answer.
Experiments on 4 LLMs across 8 tasks show that \our{} improves the causal reasoning capability with the state-of-the-art 95.33\% accuracy on CLADDER (surpassing human performance of 94.8\% for the first time) and reduces the hallucination on HaluEval with 10\% improvements.
It demonstrates that explicit causal structure modeling in LLMs can effectively mitigate logical inconsistencies in LLM outputs.
%
\end{abstract}

\begin{links}
    \link{Code, Datasets}{https://github.com/MrLYG/CDCR-SFT}
    \link{Extended version}{https://arxiv.org/abs/2508.12495}
\end{links}

\section{Introduction}

\begin{figure}[t!]
\centering
\includegraphics[width=\linewidth]{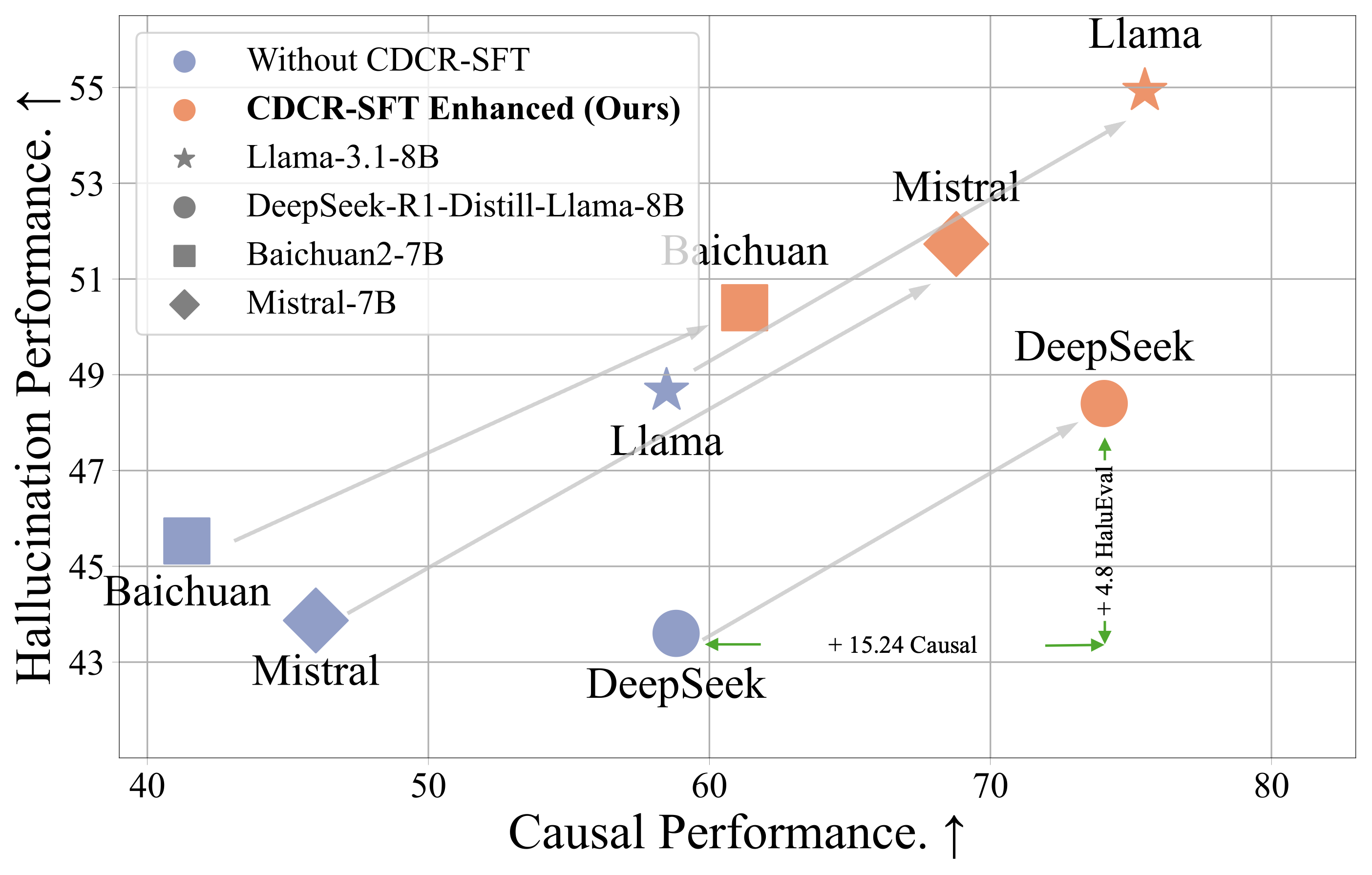}
\caption{Average overall performance of our \our{} applied to four LLMs on the \textbf{causal reasoning} benchmarks (CLADDER and WIQA) and the \textbf{hallucination} benchmark (HaluEval). Orange symbols denote the LLMs enhanced by \our{}, demonstrating that \our{} significantly improves causal reasoning capabilities and reduces hallucinations.}

\label{fig:fig1}
\end{figure}

Large language models (LLMs) may generate logically inconsistent hallucinations during reasoning, where their outputs appear coherent but contain logical inconsistencies, leading to suboptimal performance~\cite{banerjee2024llms,huang2025survey,cheng2025empowering}.
Recent studies point out a correlation between causal reasoning capabilities and these logical inconsistency hallucinations ~\cite{bagheri2024c2p,wang-2024-causalbench,liu2025large}, namely LLMs with stronger causal reasoning abilities typically exhibit fewer logically inconsistent hallucinations. 
This observation motivates the central research question of this work: ``\textit{\textbf{Can we mitigate hallucinations by improving the causal reasoning capabilities of LLMs?}}''

To answer this question, we must enhance LLMs' causal reasoning abilities.
However, true causal reasoning requires LLMs to represent and manipulate a directed acyclic graph (DAG) that encodes conditional independence relationships, enables intervention queries, and satisfies causal identification assumptions (\textit{i}.\textit{e}., exchangeability, consistency, positivity)~\cite{hernan2020causal} for identifying confounding effects.
Existing structured-reasoning methods, including Chain-of-Thought (CoT)~\cite{wei2022chain}, Tree-of-Thought (ToT)~\cite{yao2023tree}, Graph-of-Thought (GoT)~\cite{besta2024graph}, and Diagram-of-Thought (DoT)~\cite{zhang2024diagram}, operate at the wrong level of abstraction, which models dependencies between linguistic tokens rather than causal relationships between variables~\cite{bao2024likely,fu2025unveiling,luo2025causal}. 
These methods generate reasoning structures only at inference time through prompting, without any training signal to correct mis-specified causal relationships. 
Consequently, when an LLM incorrectly identifies A as causing B (when B actually causes A), or fails to recognize a confounding variable C that influences both, no gradient flows back to fix these fundamental errors~\cite{wang2023self,yao2023tree,besta2024graph}.
As a result, they cannot block spurious back-door paths or guarantee counterfactual consistency, leaving LLMs still vulnerable to logically inconsistent hallucinations~\cite{wang2023self,yao2023tree,besta2024graph}.
The mathematical constraints further compound this problem.
Causal relationships inherently form a DAG that encodes multiple interconnected variables with conditional dependencies and multiple pathways of influence.
A linear chain or even a tree structure cannot adequately represent scenarios where a variable influences multiple outcomes simultaneously or where effects depend on the interaction of multiple causes, both fundamental characteristics of causal DAG.
This structural mismatch means that prompt-only variants such as CoT, ToT, GoT, and DoT cannot, by design, supervise LLMs to learn causal edge semantics, limiting their ability to enforce conditional independencies required for true causal inference.

To address this gap, we propose \textbf{C}ausal-\textbf{D}AG \textbf{C}onstruction and \textbf{R}easoning (\our{}), a supervised fine-tuning framework that trains LLMs to first construct a variable-level causal DAG and then reason over that graph. 
The training of \our{} requires data with a causal DAG as well as the corresponding reasoning on top of that. 
Therefore, we introduce CausalDR (Causal-\textbf{D}AG and \textbf{R}easoning), the first dataset specifically designed to train LLMs in simultaneous causal DAG construction and graph-based reasoning. 
Building upon the CLADDER dataset~\cite{jin2023cladder}, which provides causal questions with a causal DAG, we develop an automated generation and validation pipeline using DeepSeek-R1~\cite{deepseekai2025}. 
This pipeline ensures high-quality data generation through question-answer consistency checks. 
Each sample in CausalDR comprises (1) an input question, (2) a causal DAG that explicitly describes variables and their relationships, (3) a graph-based reasoning trace that navigates the causal structure, and (4) the final answer.
As shown in Fig.~\ref{fig:fig1}, our experiments demonstrate that \our{} can address our research question by both improving causal reasoning capabilities and mitigating the logically inconsistent hallucinations across multiple benchmarks. 
This indicates that, rather than solely pursuing larger model sizes or more training data or longer CoT, we can achieve more trustworthy LLMs by equipping them with structured reasoning capabilities that align with the underlying causal nature of real-world problems.

The major contributions of this work are three-fold.
First, we introduce \our{}, a supervised fine-tuning framework that shifts how LLMs approach causal reasoning by moving from sequential CoT to DAG-based inference. 
It trains models to construct a causal DAG that properly encodes both causal directionality and conditional independence relationships, enabling them to perform structured reasoning over these graphs rather than being constrained by linear reasoning paths.
Second, we present CausalDR, a dataset containing 25,368 high-quality samples for teaching LLMs to generate causal DAG construction and reason on top of the DAG. 
Third, we demonstrate that explicit causal structure modeling can not only improve causal reasoning but also mitigate hallucinations in LLMs. 

\begin{figure*}[t!]
\centering
\includegraphics[width=\linewidth]{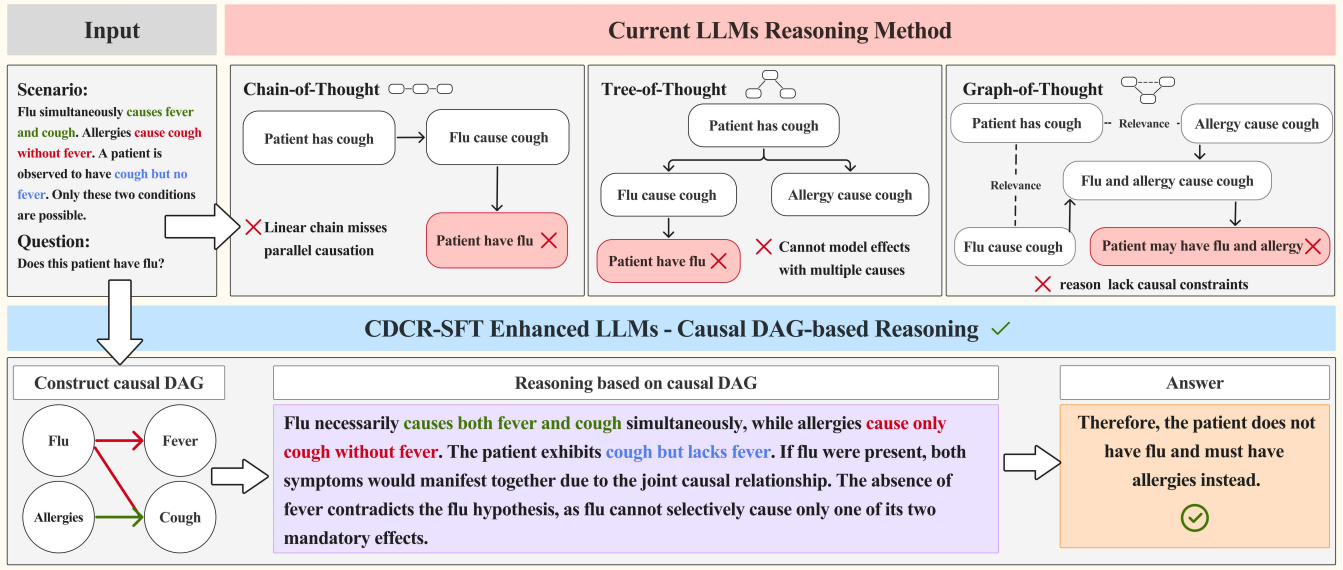}
\caption{
Comparison of reasoning approaches: Existing methods (CoT, ToT, GoT) operate at linguistic/semantic levels without explicit causal structure; Our \our{} constructs a variable-level causal DAG with directed edges representing causal relationships, enabling principled causal inference through graph-based reasoning.
}

\label{fig:cdcr_review}
\end{figure*}

\section{Related Works} 

\paragraph{Reasoning and Causal Limitations in LLMs}
LLMs employ structured reasoning methods such as Chain-of-Thought (CoT) prompting, which generates intermediate steps alongside final answers \cite{wei2022chain}; Self-Consistency (CoT-SC), which samples multiple reasoning chains for robustness; Tree-of-Thoughts (ToT), which branches into alternative solution paths \cite{yao2023tree}; and Graph-of-Thoughts (GoT), which links subproblems as nodes in a simple graph \cite{besta2024graph}. However, these methods treat inference as linear sequences or trees and cannot represent directed acyclic graph (DAG) needed for causal analysis, where edges denote cause–effect relations and support interventions and counterfactual reasoning. Benchmarks such as CausalBench show that LLMs struggle with intervention and counterfactual queries, failing to predict outcomes of hypothetical changes \cite{wang-2024-causalbench}, and synthetic tests confirm that models rely on surface text patterns rather than true cause–effect relations \cite{ma2024causal}.

\paragraph{Hallucination Reduction and Causal Supervised Fine‑Tuning}
Complex reasoning tasks can exacerbate hallucinations in LLMs, as models often rely on surface‑level correlations rather than true causal structure \cite{bagheri2024c2p}. Traditional mitigation—external knowledge checks or post‑hoc filters—only corrects errors after generation and does not strengthen the model’s internal inference process \cite{wang-2024-causalbench}. Recent studies have demonstrated that task‑specific fine‑tuning significantly improves LLM performance on specialized benchmarks \cite{han2024parameter,liu2025large}. In particular, supervised fine‑tuning (SFT) with low‑rank adapters (LoRA) \cite{hu2022lora} reshapes internal reasoning by training models on structured targets. In this study, we extend this paradigm by using the CausalDR dataset’s annotated DAG and stepwise reasoning to teach the model to first construct a causal graph and then perform graph‑based inference, thereby reducing hallucinations and improving consistency.

\section{Methods}
\subsection{\our{}}
\begin{figure}[!h]
\centering
\includegraphics[width=\linewidth]{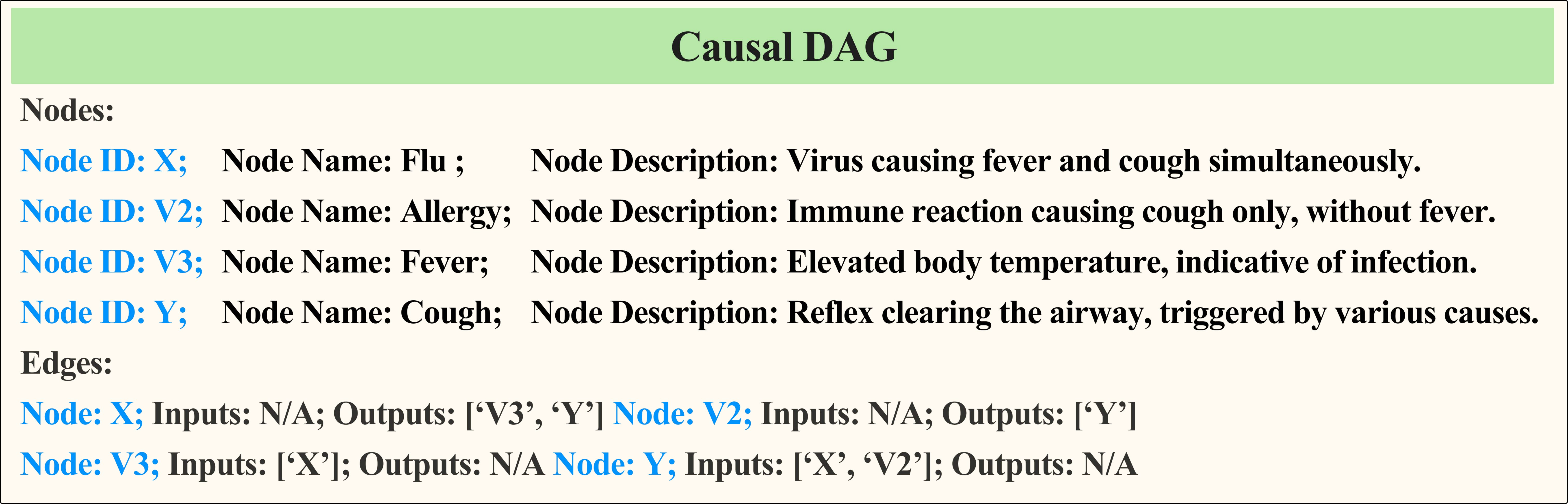}
\caption{Textual representation of the causal DAG in Fig.~\ref{fig:cdcr_review}.
}
\label{fig:textual_dag}
\end{figure}

\our{} is a supervised fine-tuning framework that trains LLMs to explicitly perform causal reasoning through Causal-DAG construction and reasoning.
Specifically, LLMs learn to construct a causal DAG by identifying causal variables from input queries, then perform structured reasoning over the DAG, and finally generate answers, as shown in Fig.~\ref{fig:cdcr_review}.
Existing structured reasoning methods, such as CoT, ToT, and GoT, generally produce reasoning paths at the linguistic token or semantic levels without modeling the underlying causal structures among variables. Table~\ref{table:cdcr_comparison} provides a detailed comparison of key capabilities between our proposed \our{} framework and existing reasoning methods.\
Mathematically, CoT generates reasoning paths as linear reasoning sequences $S_{\text{CoT}}=(p_1,\dots,p_n,y)$, ToT forms branching reasoning trees $S_{\text{ToT}}=\text{Tree}(p_1,\dots,p_n,y)$, and GoT creates semantic-level reasoning graphs $S_{\text{GoT}}=\text{Graph}(p_1,\dots,p_n,y)$. 
\begin{table}[htbp]
\centering
\small
\resizebox{0.95\linewidth}{!}{
\begin{tabular}{lcccc}
\toprule
\textbf{Aspect} & \textbf{CDCR-SFT (ours)} & \textbf{CoT} & \textbf{ToT} & \textbf{GoT}\\
\midrule
Reasoning aligned with causal relationships
 & \(\checkmark\) & \(\times\) & \(\times\) & \(\times\)\\[1ex]

Explicit causal training signal & \(\checkmark\) & \(\times\) & \(\times\) & \(\times\)\\[1ex]

Supports multi-parent causes & \(\checkmark\) & \(\times\) & \(\times\) & \(\times\) (no causal)\\[1ex]

Captures conditional independencies & \(\checkmark\) & \(\times\) & \(\times\) & \(\times\)\\[1ex]

Captures interventions & \(\checkmark\) & \(\times\) & \(\times\) & \(\times\)\\[1ex]

Captures counterfactuals & \(\checkmark\) & \(\times\) & \(\times\) & \(\times\)\\[1ex]

Effective hallucination mitigation & \(\checkmark\) & \(\times\) & \(\times\) & \(\times\)\\
\bottomrule
\end{tabular}
}
\caption{Comparison of key capabilities between \our{} and existing reasoning methods.}
\label{table:cdcr_comparison}
\end{table}
\our{} outputs a DAG-based reasoning process $S_{\our{}}=(G, P, y)$, where $G=(V, E)$ denotes the causal DAG encoding causal directionality and conditional independence relationships, $P=(p_1(G),\dots,p_n(G))$ represents reasoning steps that adhere strictly to causal structures in $G$, and $y$ is the final inferred answer.
Specifically, in the textual encoding of the causal DAG G, each causal variable is clearly represented as a node described in natural language, including detailed descriptions of the primary events. The causal relationships among these variables are encoded as directed edges, explicitly indicating directional influences.
An illustrative example of textual DAG encoding is provided in Fig.~\ref{fig:textual_dag}.

\subsection{Dataset Construction}

\subsubsection{Causal-DAG and Reasoning (CausalDR) Dataset}
\label{section:causal_dag_reasoning}
To train LLMs in simultaneous causal DAG construction and graph-based reasoning, we require datasets explicitly providing supervision for both. However, existing causal datasets~\cite{gordon-etal-2012-semeval,tandon2019wiqa,du2022care} either omit explicit causal relationships altogether or, as exemplified by CLADDER~\cite{jin2023cladder}, offer mathematically rigorous yet semantically sparse causal graphs and algebraic formulations, lacking clear natural-language reasoning paths linking structures to answers (A CLADDER example is provided in Appx.~A.1).

We introduce \textbf{CausalDR}, the first large-scale annotated dataset explicitly designed for supervised fine-tuning of LLMs in simultaneous causal DAG construction and structured causal reasoning. Each training sample in CausalDR consists of: (1) an instruction specifying the task, (2) an input question or scenario, and (3) a coherent output comprising three components: a text-based causal DAG $G$, a reasoning path $P==(p_1(G),\dots,p_n(G))$ based on $G$, and a final answer $y$ derived through structured inference (A detailed example see Appx.~A.2). 
\begin{figure}[h!]
\centering
\includegraphics[width=0.95\linewidth]{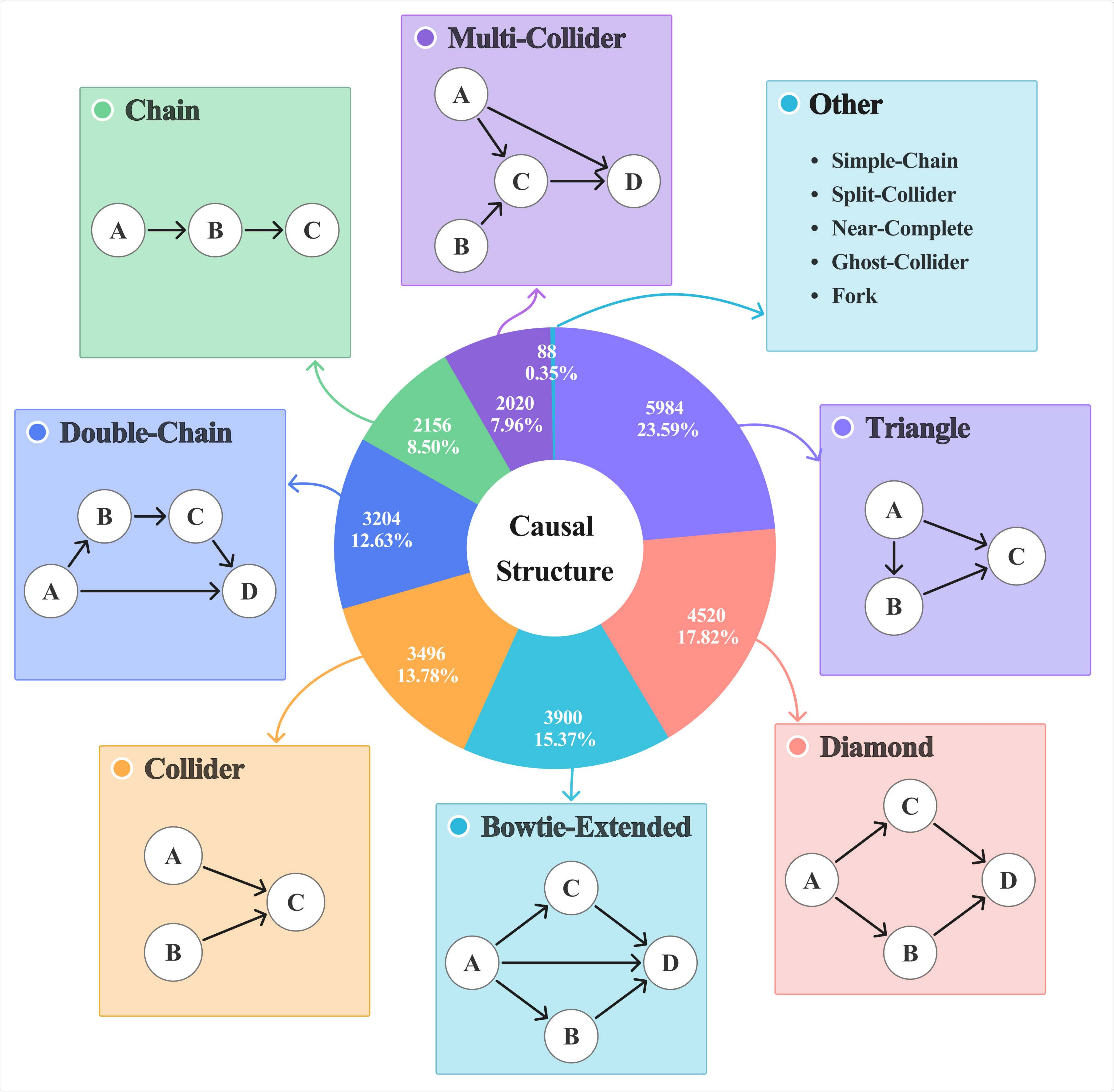}
\caption{Proportional Distribution of 12 Canonical Causal DAG Structures in the CausalDR Dataset.}
\label{fig:causaldr_structure_distribution}
\end{figure}
We construct CausalDR based on the CLADDER dataset~\cite{jin2023cladder}, partitioning it into training and test sets based on unique identifiers (\texttt{graph\_id} and \texttt{story\_id}) to prevent information leakage. 
And then using the DeepSeek-R1~\cite{deepseekai2025} (temperature=0.6, max tokens=8192, details in Appx.~A.3), we developed an automated pipeline (pseudocode provided in Appx.~A.5) to generate and validate high-quality training samples for the CausalDR dataset.
Specifically, we designed a prompt (details in Appx.~A.4) that contains a mathematically accurate causal DAG expressed in formal notation, instructing DeepSeek-R1 to produce JSON-formatted outputs for each CLADDER sample. 
Each output explicitly: (1) causal nodes with clear semantic descriptions, and causal edges specifying incoming and outgoing relationships, (2) a step-by-step reasoning path that explicitly references the constructed causal DAG, and (3) the final inferred answer. 
To ensure quality, we implemented a validation mechanism comparing model-generated answers against the original ground-truth answers provided by CLADDER. If a generated answer did not match the ground-truth after multiple validation attempts, the sample was manually reviewed or discarded. Through this process, we obtained a high-quality dataset of 6,357 validated samples. 

To further enhance dataset diversity and generalization, we introduced a Causal DAG Augmentation technique. 
Specifically, given an original causal DAG $G=(V,E)$, we randomly permuted the order of causal nodes and edges using permutation functions $\pi_v(\cdot)$ and $\pi_e(\cdot)$, respectively, to create diverse augmented variants:
$V_{\text{aug}}=\pi_v(V),\quad E_{\text{aug}}=\pi_e(E),\quad G_{\text{aug}}=(V_{\text{aug}},E_{\text{aug}})$.
We applied this permutation procedure four times per original DAG $G$, each time pairing the permuted DAG $G_{\text{aug}}$ with the original reasoning path $P$ and answer $y$. This expanded the initial dataset from 6,357 samples to 25,368 augmented training examples. 

Fig.~\ref{fig:causaldr_structure_distribution} shows the proportional distribution of the 12 canonical causal DAG structures within the CausalDR dataset. These structures cover diverse causal configurations, including simple Chains (e.g., Chain, Double-Chain), Confounding structures (e.g., Triangle, Fork), Colliders (e.g., Collider, Multi-Collider), and more intricate multi-path interactions (e.g., Diamond, Bowtie-Extended). The diverse representation of these key causal mechanisms enables effective generalization of causal reasoning capabilities in large language models.

\subsubsection{Auxiliary Instruction following Data}
To prevent the model from over-focusing on the causal task and degrading the linguistic generalization ability, we randomly select
10,000 Alpaca~\cite{alpaca} examples and mix them with the CausalDR dataset during supervised fine‑tuning to ensure the overall linguistic ability and generalization performance of the model.

\subsection{Supervised Fine-tuning Procedure}
\label{sec:sft-procedure}

During supervised fine-tuning, LLM learns to generate the structured causal DAG inference sequence $S_{\our{}}=(G, P, y)$. The optimization objective is formulated as a negative log-likelihood loss:
$
\mathcal{L}_{\textsc{\our{}}} = -\sum_{t=1}^{|S|}\log P(s_t \mid s_{<t}, X)
$
, where $s_t$ denotes the $t$-th token in the ground-truth sequence $S$, and $s_{<t}$ represents all tokens before position $t$. 

Critically, whenever the model-generated sequences deviate from the ground-truth causal DAG structure—such as introducing reversed causal edges, omitting essential causal variables, or adding extraneous causal relationships—explicit gradient signals immediately correct these inaccuracies. This supervision ensures that the model internalizes correct causal directionality, conditional independence properties, and intervention semantics required for accurate causal reasoning.
For computational efficiency, we applied Low-Rank Adaptation (LoRA)~\cite{hu2022lora} during fine-tuning, updating only a small number of low-rank parameters inserted into each layer, while freezing the original pretrained LLM parameters. 
Through this fine-tuning procedure, \our{} trains the model to construct an accurate causal DAG and perform structured reasoning explicitly constrained by the causal relationships defined in these graphs, thereby improving the logical consistency of the LLM outputs and mitigating hallucinations.

\section{Experiments}
\label{sec:experiments}

\subsection{Experimental Setup}

\subsubsection{Base LLMs and Reasoning Methods}
We select 4 pretrained LLMs for evaluation: (1) Llama-3.1-8B-Instruct~\cite{grattafiori2024llama}, (2) DeepSeek-R1-Distill-Llama-8B~\cite{deepseekai2025}, (3) Baichuan2-7B-Chat~\cite{baichuan2023baichuan2}, and (4) Mistral-7B-Instruct-v0.2~\cite{jiang2023mistral7b}. 
We compare our \our{} method against 5 baseline reasoning approaches: Zero-shot-CoT(\textbf{CoT})\cite{kojima2023largelanguagemodelszeroshot}, Chain-of-Thought Self-Consistency (\textbf{CoT-SC})\cite{wang2023self}, Causal Chain-of-Thought (\textbf{CausalCoT})\cite{jin2023cladder}, Tree-of-Thoughts (\textbf{ToT})\cite{yao2023tree}, and Graph-of-Thoughts (\textbf{GoT})~\cite{besta2024graph}. Detailed descriptions of each baseline are provided in Appx.~B.1.
Additionally, DeepSeek-R1-Distill-Llama-8B, which is based on Llama-3.1-8B and fine-tuned on high-quality reasoning data~\cite{deepseekai2025}, serves as a supervised fine-tuning reasoning baseline.

\subsubsection{Datasets}
We conduct experiments on 3 distinct datasets (Cladder~\cite{jin2023cladder}, WIQA~\cite{tandon2019wiqa}, and HaluEval~\cite{HaluEval}) to evaluate models’ causal reasoning and hallucinations performance.

\begin{table}[ht!]
\resizebox{\linewidth}{!}{
\begin{tabular}{lccc}
\toprule
\textbf{Dataset} & \textbf{\#} & \textbf{Subtasks} & \textbf{Evaluation Focus} \\ 
\midrule
CLADDER & 1,906 & Rung 1, Rung 2, Rung 3 & Causal reasoning; Causal DAG quality \\ 
\midrule
WIQA &  212 & INPARA, EXOGENOUS & Causal reasoning\\ 
\midrule
HaluEval  & 1,500 & Dialogue, QA, Summarization & Hallucination \\ 
\bottomrule
\end{tabular}
}
\caption{Summary of datasets used in experiments.}
\end{table}

\textbf{CLADDER}~\cite{jin2023cladder}: 
A benchmark dataset evaluating LLMs’ causal reasoning at three levels: Rung 1 (Association, observational correlations), Rung 2 (Intervention, active manipulation effects), and Rung 3 (Counterfactual, hypothetical “what-if” scenarios). Following preprocessing (see section~\ref{section:causal_dag_reasoning}), CLADDER is split into training and test sets by \texttt{graph\_id} and \texttt{story\_id} to avoid data leakage. To further ensure test data quality, we perform an additional validation step (details in Appx.~B.2).
\textbf{WIQA}~\cite{tandon2019wiqa}: A challenging dataset for evaluating LLMs’ causal reasoning capabilities. We focus on two perturbation types: in-paragraph (INPARA), which changes within the text that test causal chain reconstruction, and out-of-paragraph (EXOGENOUS), which external changes assess the model’s reasoning about external influences (more information see Appx.~B.3). 
\textbf{HaluEval}~\cite{HaluEval}: A benchmark for evaluating models’ hallucination across three NLP tasks: (1) Knowledge-grounded Dialogue (Dialogue), (2) Question Answering (QA), and (3) Text Summarization (Summarization). Each task includes paired examples, consisting of hallucinated samples (incorrect or unverifiable information) and corresponding factual samples. For our experiments, we randomly sample 500 pairs per task (total 1,500 pairs).

\begin{table*}[!t]
\centering
\resizebox{\textwidth}{!}{
\begin{tabular}{lcccc|ccc|cccc}
\toprule
\toprule
\multicolumn{1}{c}{\multirow{2}{*}{\textbf{Method}}} & \multicolumn{4}{c}{\textbf{Cladder (\%)$\uparrow$}} & \multicolumn{3}{c}{\textbf{WIQA (\%)$\uparrow$}} & \multicolumn{4}{c}{\textbf{HaluEval (\%)$\uparrow$ }} \\ \cline{2-12} 
\multicolumn{1}{c}{} & Rung1 & Rung2 & Rung3 &\textit{overall.} & INPARA & EXOGENOUS &\textit{overall.} & Dialogue & QA & Summarization &\textit{overall.}\\ 
\midrule
\multicolumn{12}{c}{\textbf{Llama-3.1-8B}} \\
CausalCoT & 70.90 & 72.82 & 57.46 & 65.90 & 48.11 & 33.96& 41.04 & 56.40 & 42.60 & 56.20 & 51.73  \\
CoT & 69.07 & 82.06 & 57.33 & 66.95  & 54.72 & 45.28 & 50.00  & 50.60 & 39.80 & 55.60 & 48.67 \\
CoT-SC & 72.87 & 88.13 &  65.31 & 72.88  & 60.38 & 44.34 & 52.36  & 43.60 & 34.00 & 52.60 & 43.40 \\                                              
ToT & 71.17 & 79.16 &  64.79 & 70.20 & 56.60 & 45.28  & 50.94 & 52.80 & 42.00 & 58.00 & 50.93 \\
GoT & 61.21 & 76.78 &  58.90 & 63.38 & 55.66 & 47.17  & 51.42 & 50.20 & 43.40 & 50.20 & 47.93 \\
\textbf{\our{} (Ours)} & \textbf{98.30} & \textbf{93.93} &  \textbf{93.06} & \textbf{95.33} & \textbf{64.20} & \textbf{47.20} & \textbf{55.66} & \textbf{60.80} & \textbf{44.80} & \textbf{59.20} & \textbf{54.93} \\
\midrule
\multicolumn{12}{c}{\textbf{DeepSeek-R1-Distill-Llama-8B}} \\
CausalCoT & 74.97 & 68.87 &  59.03 & 67.37 & 52.83 & 50.94  & 51.89 & 47.60 & 40.00 & 51.80 & 46.47 \\              
CoT & 73.92 & 76.78 &  53.27 & 66.21 & 55.66 & 47.17  & 51.42 & 42.00 & 40.80 & 48.00 & 43.60 \\          
CoT-SC & 77.98 & 88.13 &  63.74 & 74.29 & 51.89 & 43.40  & 47.64 & 33.60 & 41.40 & 32.20 & 35.73 \\
ToT & 70.34 & 80.62 &  66.18 & 71.14 & 56.60 & 44.34  & 50.47 & 39.40 & 43.80 & 40.20 & 41.13 \\  
GoT & 75.23 & 80.97 &  57.63  & 69.43 & 55.66 & 50.00  & 52.83 & \textbf{53.40}  & 41.00 & 50.60 & 48.33 \\         
\textbf{\our{} (Ours)} & \textbf{94.89} & \textbf{90.50} &  \textbf{90.97} & \textbf{92.44} & \textbf{56.60} & \textbf{54.72} & \textbf{55.66}& 48.60 & \textbf{44.40} & \textbf{52.60} & \textbf{48.53} \\
\midrule
\multicolumn{12}{c}{\textbf{Baichuan2-7B}} \\
CausalCoT & 50.46 & 46.44 & 51.70 & 50.16 & 22.64 & 27.36  & 25.00 & 45.80 & 48.40 & 43.20 & 45.80 \\               
CoT & 49.67 & 62.01 &  48.56 & 51.68 & 34.91  & 27.36  & 31.13 & 44.20 & 46.60 & 45.80 & 45.53 \\
CoT-SC & 51.38 & 61.21 &  48.69 & 52.26 & 36.79 & 30.19  & 33.49 & 47.80 & 45.80 & 47.40 & 47.00 \\                                     
ToT & 49.67 & 58.05 &  50.65 & 51.73 & 34.91 & 20.75  & 27.83 & 44.80  & 45.80 & 48.01 & 46.20 \\
GoT & 51.11 & 58.84 &  49.61 & 52.05 & 31.13 & 30.19  & 30.66 & 41.80 & 43.80 & 40.80 & 42.13 \\
\textbf{\our{} (Ours)} & \textbf{71.04} & \textbf{75.20} &  \textbf{72.64} & \textbf{72.51} & \textbf{50.00} & \textbf{50.00} & \textbf{50.00} & \textbf{50.60} & \textbf{49.60} & \textbf{51.00} & \textbf{50.40} \\
\midrule
\multicolumn{12}{c}{\textbf{Mistral-7B}} \\
CausalCoT & 51.11 & 63.06 &  45.16 & 51.10 & 38.68  & 27.36  & 33.02 & 45.20 & 47.80 & 41.60 & 44.87 \\                                               
CoT & 52.29 & 59.63 &  53.53 & 54.25 & 40.57  & 34.91 & 37.74 & 43.60 & 44.20 & 43.80 & 43.87 \\  
CoT-SC & 56.75 & 66.75 &  58.90 & 59.60 & 42.45 & 38.68  & 40.57 & 44.40 & 45.20 & 44.00 & 44.53 \\                                               
ToT & 50.46 & 56.20 &  50.39 & 51.57 & 42.45 & 32.08 & 37.26 & 47.00 & 42.80 & 46.60 & 45.47 \\
GoT & 50.85 & 63.85 &  56.15 & 55.56 & 42.45 & 40.57 & 41.51  & 47.60 & 46.20 & 46.80 & 46.87 \\
\textbf{\our{} (Ours)} & \textbf{94.23} & \textbf{94.46} &  \textbf{90.45} & \textbf{92.76} & \textbf{43.40} & \textbf{46.23} & \textbf{44.81} & \textbf{53.40} & \textbf{48.20} & \textbf{53.60} & \textbf{51.73} \\
\bottomrule
\bottomrule
\end{tabular}
}
\caption{Performance comparison between our proposed \our{} and baseline reasoning methods on causal reasoning benchmarks (CLADDER and WIQA) and hallucination benchmark (HaluEval) across four different LLMs. Accuracy (\%) is reported for overall benchmarks and subtasks; best results per model and task highlighted in bold.
}

\label{table:main_results}
\end{table*}

\subsubsection{Evaluation Metrics}
We adopt two primary metrics to clearly evaluate the models’ causal reasoning and hallucination reduction: (1) \textbf{Accuracy}: measures correctness in causal reasoning (CLADDER, WIQA) and hallucination (HaluEval) tasks. (2) \textbf{Causal DAG Quality}: evaluates \textit{Node Score} (correct causal nodes), \textit{Edge Score} (correct causal edges), and \textit{Structural Score} (overall graph correctness, including directionality and completeness). Causal DAG is scored using GPT-4o-mini~\cite{hurst2024gpt}, with detailed scoring criteria and evaluation procedures provided in Appx.~B.4.

\subsubsection{Implementation Details}
We perform LoRA fine-tuning on A40x4 GPUs using the LLaMA-Factory library~\cite{zheng2024llamafactory} with default hyperparameters. Fine-tuned models use vLLM~\cite{kwon2023efficient} on the same GPUs for inference. Base model inference is conducted through external platforms: DeepInfra for Llama-3.1-8B and Mistral-7B, and Baidu-Qianfan/OpenRouter for Baichuan2-7B and DeepSeek-R1-Distill-Llama-8B, with 200 concurrent threads. The inference temperature is set to 0.0 except for DeepSeek (0.6, following~\cite{deepseekai2025}), CoT-SC (0.7, following~\cite{wang2023self}), and GoT (1.0, following~\cite{besta2024graph}). Our method and CoT-based approaches utilize a unified three-step instruction, while CausalCoT, ToT, and GoT follow their original structured prompting\cite{jin2023cladder, yao2023tree, besta2024graph}. All reported results are averaged over 3 experimental runs. Complete implementation details, prompts, and configurations are provided in our available code.

\subsection{Main Results and Analysis}
\label{sec:main_results}

\noindent
\textbf{Causal Reasoning Performance.} 
\label{sec:causal_reasoning_performance}
Table~\ref{table:main_results} reports the causal reasoning performance of our proposed \our{} method compared with five baseline methods (CoT, CoT-SC, CausalCoT, ToT, and GoT) across four different LLMs on two representative causal reasoning benchmarks: CLADDER and WIQA.

On the CLADDER benchmark, our \our{} consistently achieves improvements across all three causal reasoning levels (Rung 1: Association, Rung 2: Intervention, and Rung 3: Counterfactual). Specifically, with the Llama-3.1-8B-Instruct model, our method reaches an overall accuracy of 95.33\%, surpassing the strongest baseline (CoT-SC: 72.88\%) by an absolute margin of 22.45 percentage points. Remarkably, at the most challenging Counterfactual reasoning level (Rung 3), \our{} achieves a particularly large improvement of 27.75 percentage points, improving accuracy from 65.31\% (CoT-SC) to 93.06\%. More importantly, our approach is the first to surpass the human-level benchmark performance (94.8\%)~\cite{yu2025causalevalbettercausalreasoning} on CLADDER. Similar consistent performance gains are also observed for the DeepSeek-R1-Distill-Llama-8B (74.29\% to 92.44\%), Baichuan2-7B-Chat (52.26\% to 72.51\%), and Mistral-7B-Instruct-v0.2 (59.60\% to 92.76\%) models.

On the WIQA benchmark, \our{} again achieves consistent improvements over all baseline methods. Taking the Llama-3.1-8B-Instruct model as an example, the overall accuracy is improved from the best baseline (CoT-SC: 52.36\%) to 55.66\%. Similar improvements are consistently observed for the DeepSeek-R1-Distill-Llama-8B (52.83\% to 55.66\%), Baichuan2-7B-Chat (33.49\% to 50.00\%), and Mistral-7B-Instruct-v0.2 (41.51\ to 44.81\%) models.

These consistent gains across multiple causal reasoning tasks (CLADDER and WIQA) and diverse model architectures—from instruction-tuned models (Llama-3.1-8B-Instruct and Mistral-7B-Instruct-v0.2) to distilled variants (DeepSeek-R1-Distill-Llama-8B) and smaller-scale models (Baichuan2-7B-Chat)—reflect that the benefits of \our{} originate primarily from its explicit modeling of causal structures reasoning. Unlike conventional methods that perform token-level or semantic-level reasoning, our approach trains LLMs to explicitly construct and reason over causal DAG, thus embedding a stronger inductive bias aligned with causal inference principles. Consequently, the models internalize improved representations of conditional independencies, intervention semantics, and causal directionality, facilitating more robust generalization across causal reasoning scenarios and tasks of varying complexity.

\noindent
\textbf{Hallucination Reduction.} 
Table~\ref{table:main_results} further reports the hallucination reduction performance of our proposed \our{} method across four different LLMs, evaluated on the HaluEval benchmark comprising three typical tasks: Dialogue, QA, and Summarization.

Our \our{} method consistently outperforms baseline reasoning methods in terms of overall accuracy on the HaluEval benchmark, demonstrating clear reductions in logical inconsistencies and hallucinations. 
Specifically, using the Llama-3.1-8B model, \our{} achieves an overall accuracy of 54.93\%, significantly higher than the strongest baseline (CausalCoT: 51.73\%) and substantially surpassing CoT-SC (43.40\%) by over 11 percentage points. Particularly noteworthy is the Dialogue subtask, where accuracy improves from 43.60\% (CoT-SC) to 60.80\%, highlighting the effectiveness of our approach in mitigating hallucinations in complex interactive reasoning tasks.

Similar trends are evident for other evaluated LLMs. For instance, the DeepSeek improves from the strongest baseline (CausalCoT: 46.47\%) to 48.40\%, Baichuan improves from 47.00\% (CoT-SC) to 50.40\%, and Mistral shows accuracy improvement from the best baseline (GoT: 46.87\%) to 51.73\%. Importantly, these significant hallucination reductions are achieved without hallucination-focused supervision, indicating that the reduction naturally arises from enhanced causal reasoning capabilities learned by the model.

\noindent
\textbf{SFT-based comparison.} 
DeepSeek-R1-Distill-Llama-8B serves as an SFT-trained reasoning baseline. Under the same Llama-8B, CDCR-SFT (Llama-3.1-8B) outperforms it on WIQA (55.66\% vs. 51.42\%) and HaluEval (54.93\% vs. 43.60\%), showing stronger causal reasoning and lower hallucination.

These empirical findings directly support our core hypothesis: explicitly improving the causal reasoning capabilities of LLMs inherently mitigates logically inconsistent hallucinations. The substantial and consistent hallucination reductions observed across diverse tasks and model architectures demonstrate that our \our{} method provides an effective and generalizable solution for enhancing the reliability and consistency of LLMs.

\subsection{Causal DAG Construction Quality}
\our{} is to enable LLMs to reason accurately based on a variable-level causal DAG. The quality of the generated DAG thus directly reflects the extent to which the model has internalized correct causal relationships and structured causal reasoning capabilities, including accurately capturing causal directionality, conditional independencies, and satisfying causal identification assumptions.
\begin{figure}[h!]
\centering
\includegraphics[width=\linewidth]{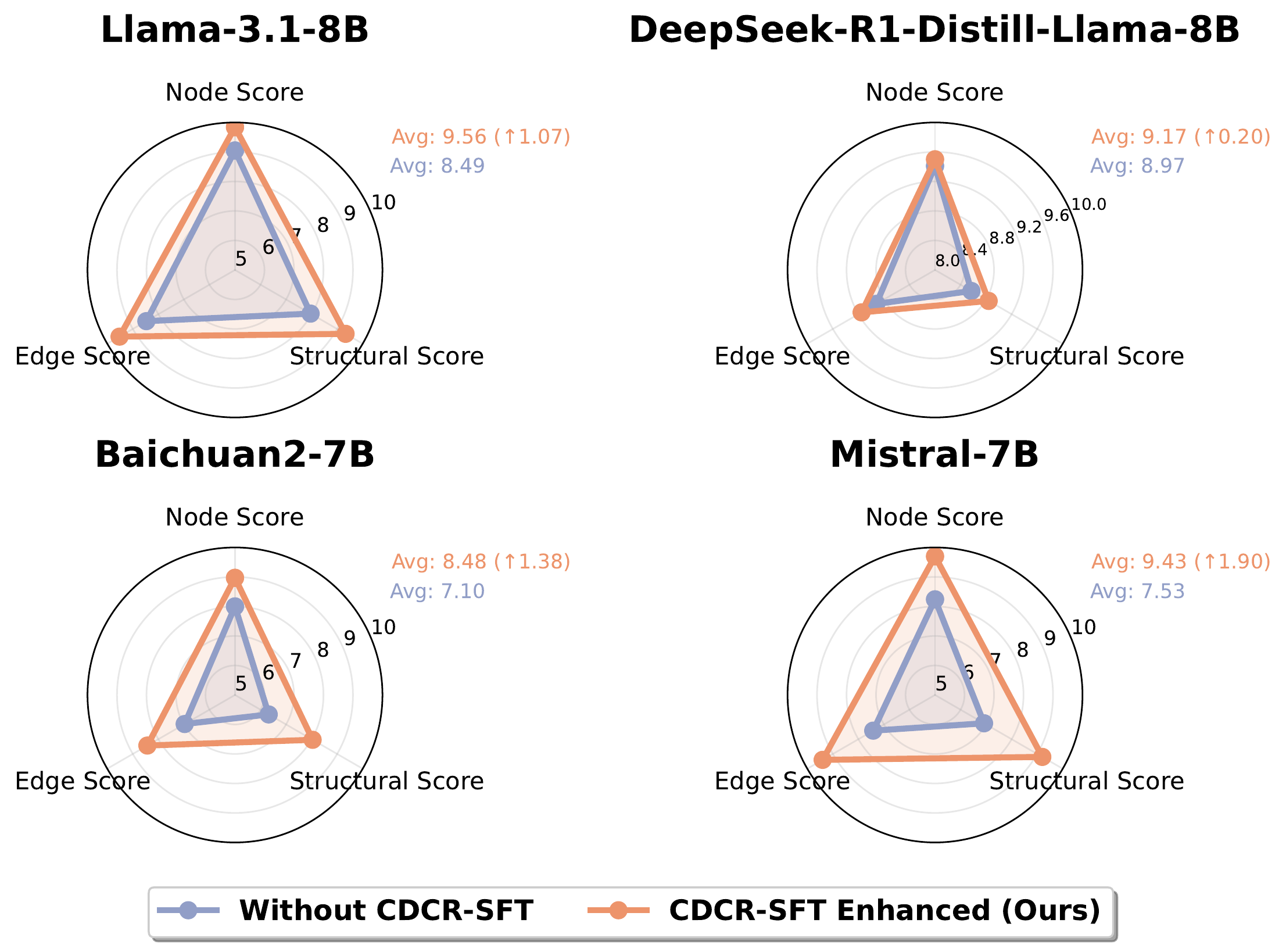}
\caption{Comparison of causal DAG quality scores (Node, Edge, and Structural Scores) generated by pretrained LLMs versus those enhanced with \our{}, evaluated on the CLADDER dataset.}

\label{fig:causal_graph_quality_radar}
\end{figure}
We compare the Causal DAG generated using pre-trained LLMs versus the DAG produced by LLMs enhanced with our \our{} approach. Both employ the same prompt instructing models to generate Causal DAG. Fig.~\ref{fig:causal_graph_quality_radar} indicates that \our{} raises scores in each dimension for all models. For Llama-3.1-8B, the overall average increases from 8.49 to 9.56, with the largest rise in Structural Score (7.96 to 9.33). DeepSeek-R1-Distill-Llama-8B shows a small increase from 8.97 to 9.17, chiefly in Edge Score (8.92 to 9.15). Baichuan2-7B advances from 7.10 to 8.48, with a 1.72-point gain in Structural Score. Mistral-7B displays the greatest progress, from 7.53 to 9.43, with gains over 2 points in both Edge Score and Structural Score.
The significantly higher DAG quality achieved by \our{} over baseline methods validates that explicit DAG-based reasoning supervision enhances LLMs’ capability to correctly represent and reason with causal structures, directly supporting improvements observed in causal reasoning tasks and hallucination reduction.


\subsection{Ablation Study}
We conduct an ablation study to assess whether the observed performance improvements originate specifically from our causal DAG construction and causal DAG-based reasoning strategy, or merely from the additional exposure to causal knowledge and examples provided during fine-tuning. 
Specifically, we compare three experimental conditions across all three benchmarks, reporting overall accuracy for CLADDER, WIQA, and HaluEval: (i)~\textit{Baseline}: the best-performing existing reasoning method per benchmark (selected from CoT, CoT-SC, ToT, GoT, and CausalCoT in Table~\ref{table:main_results}); (ii)~\textit{\our{}-Ablated}: fine-tunes LLMs using only question-answer pairs from the CausalDR dataset, omitting causal DAG $G$ construction and reasoning paths $P$, but retaining identical auxiliary instruction-following data; (iii)~\textit{\our{}}: our full proposed method, explicitly trained on causal DAG construction and DAG-based reasoning. All conditions maintain identical training configurations, including model architectures, hyperparameters, and data volumes, ensuring a fair comparison.
\begin{table}[!h]
\centering
\resizebox{\linewidth}{!}{
\begin{tabular}{lccc}
\toprule
\toprule
\multicolumn{1}{c}{\textbf{Method}} & \multicolumn{1}{c}{\textbf{Cladder (\%)$\uparrow$}} & \multicolumn{1}{c}{\textbf{WIQA (\%)$\uparrow$}} & \multicolumn{1}{c}{\textbf{HaluEval (\%)$\uparrow$ }} \\ 
\midrule
\multicolumn{4}{c}{\textbf{Llama-3.1-8B}} \\
Baseline & 72.88 & 52.36 &  51.73 \\
\our{}-Ablated & 87.25 & 49.06 &  44.97 \\
\textbf{\our{} (Ours)} & \textbf{95.33} & \textbf{55.66} &  \textbf{54.93} \\
\midrule
\multicolumn{4}{c}{\textbf{DeepSeek-R1-Distill-Llama-8B}} \\
Baseline & 74.29 & 52.83 &  48.33 \\        
\our{}-Ablated & 74.87 & 51.89 &  43.67 \\
\textbf{\our{} (Ours)} & \textbf{92.44} & \textbf{55.66} &  \textbf{48.53} \\
\midrule
\multicolumn{4}{c}{\textbf{Baichuan2-7B}} \\
            
Baseline & 52.26 & 33.49 &  47.00 \\
\our{}-Ablated & 69.57 & 42.92 &  42.10 \\
\textbf{\our{} (Ours)} & \textbf{72.51} & \textbf{50.00}  &  \textbf{50.40} \\
\midrule
\multicolumn{4}{c}{\textbf{Mistral-7B}} \\

Baseline & 59.60 & 41.51 &  46.87 \\ 
\our{}-Ablated & 67.58 & 38.68 &  49.10 \\
\textbf{\our{} (Ours)} & \textbf{92.76} & \textbf{44.81} &  \textbf{51.73} \\
\bottomrule
\bottomrule
\end{tabular}
}
\caption{Ablation study verifying the impact of explicit causal DAG-based reasoning, comparing baseline (best existing method), \our{}-Ablated (fine-tuned without causal DAG construction and reasoning), and our \our{} across three benchmarks on four LLMs.
}
\label{table:ablation}
\end{table}
Table~\ref{table:ablation} shows that fine-tuning models solely with causal question-answer pairs (\our{}-Ablated), without explicit causal DAG-based reasoning, consistently improves accuracy on the CLADDER benchmark (e.g., +14.4\% on Llama-3.1-8B, +17.3\% on Baichuan2-7B) but leads to performance degradation on the WIQA and HaluEval benchmarks compared to the Baseline. In contrast, our full method (\our{}), which learned causal DAG construction and causal DAG-based reasoning, consistently outperforms both the Baseline and \our{}-Ablated methods across all benchmarks and model architectures. These results confirm that the observed performance gains are attributable to structured causal reasoning rather than simply additional causal data exposure.

\section{Conclusion}

We propose \our{}, to shift how LLMs approach causal reasoning by moving from sequential CoT or graph variant to causal DAG-based reasoning. It
trains models to construct a causal DAG that properly encodes both causal directionality and conditional independence relationships, enabling them to perform structured reasoning over the graph rather than being constrained by linear reasoning paths or causal-irrelevant graph reasoning. And we create the CausalDR dataset, containing 25,368 validated samples, provides high-quality supervision for LLMs to learn explicit causal DAG construction and graph-based reasoning. 
Our experiments across four LLMs on the CLADDER, WIQA, and HaluEval benchmarks demonstrate that \our{} significantly improves causal reasoning, achieving a state-of-the-art accuracy of 95.33\% on CLADDER, surpassing human performance (94.8\%) for the first time. Moreover, \our{} reduces  hallucination in HaluEval by up to 11\%, confirming that enhanced causal reasoning directly mitigates hallucinations.
These results affirmatively answer our research question: \textbf{improving the causal reasoning capabilities of LLMs can mitigate hallucinations}.
In the future,  rather than solely pursuing larger model sizes or more training data or longer CoT, we
can achieve more trustworthy LLMs by equipping them with structured reasoning capabilities that align with the underlying causal nature of real-world problems.

\section*{Acknowledgments}
This work was partially supported by the National Science Foundation under Award No.~2428039, No.~2346158, and No.~2449280, Capital One Research Awards, and Amazon Research Awards.
We also acknowledge the use of computational resources provided by the Advanced Cyberinfrastructure Coordination Ecosystem \cite{boerner2023access}: Services \& Support (ACCESS) program, supported by NSF grants \#2138259, \#2138286, \#2138307, \#2137603, and \#2138296. Specifically, this work used NCSA Delta GPU at the National Center for Supercomputing Applications (NCSA) through allocation CIS250073.  
Any opinions, findings, conclusions, or recommendations expressed in this material are those of the authors and do not necessarily reflect the views of the National Science Foundation, Capital One, and Amazon.

\bibliography{aaai2026}
\appendix
\input{appendix}
\end{document}

%% file: appendix.tex
\onecolumn 
\section*{Supplementary Material}

\section{Additional Methodological Details}
\label{app:method-details}

\subsection{Example of the Cladder Dataset}
\label{app:cladder-example}

Our CausalDR dataset for supervised fine-tuning is based on the publicly available dataset, Cladder. The Cladder dataset was initially proposed to evaluate causal reasoning capabilities of large language models (LLMs). The dataset provides scenarios, questions, formal symbolic reasoning steps, and answers. Here, we showcase only the attributes directly relevant to our data construction process.

Below is a representative example from the Cladder dataset:

\begin{tcolorbox}[colback=gray!5!white, colframe=gray!75!black, title=Original Cladder Dataset Sample]
\textbf{Scenario \& Question:} Imagine a self-contained, hypothetical world with only the following conditions, and without any unmentioned factors or causal relationships: Husband has a direct effect on wife and alarm clock. Wife has a direct effect on alarm clock. For husbands that don't set the alarm and wives that don't set the alarm, the probability of ringing alarm is 8\%. For husbands that don't set the alarm and wives that set the alarm, the probability of ringing alarm is 54\%. For husbands that set the alarm and wives that don't set the alarm, the probability of ringing alarm is 41\%. For husbands that set the alarm and wives that set the alarm, the probability of ringing alarm is 86\%. For husbands that don't set the alarm, the probability of alarm set by wife is 74\%. For husbands that set the alarm, the probability of alarm set by wife is 24\%. If we disregard the mediation effect through wife, would husband positively affect alarm clock?

\textbf{Reasoning:} 

Let X = husband; V2 = wife; Y = alarm clock. 

X -\textgreater V2, X -\textgreater Y, V2 -\textgreater Y

$E[Y_{X=1, V2=0} - Y_{X=0, V2=0}]$

$\sum_{V2=v} P(V2=v|X=0)*[P(Y=1|X=1,V2=v) - P(Y=1|X=0, V2=v)]$

$P(Y=1 | X=0, V2=0) = 0.08$

$P(Y=1 | X=0, V2=1) = 0.54$

$P(Y=1 | X=1, V2=0) = 0.41$

$P(Y=1 | X=1, V2=1) = 0.86$

$P(V2=1 | X=0) = 0.74$

$P(V2=1 | X=1) = 0.24$

$0.74 * (0.86 - 0.41) + 0.24 * (0.54 - 0.08) = 0.32$

$0.32 > 0$

\textbf{Answer:} Yes
\end{tcolorbox}
In this example, the scenario describes causal interactions among variables, the question focuses explicitly on causal effects, and the reasoning provides symbolic and probabilistic calculations. These symbolic representations serve as important references when identifying causal nodes and relationships for our structured data generation.

\subsection{Example of the CausalDR Dataset(Ours)}
\label{app:cdr-data-structure}
CausalDR dataset used for supervised fine-tuning within the \our{} framework is presented in a clear, structured JSON format. Each sample contains three essential components: (1) a specific \textbf{instruction} explicitly guiding the model on what kind of task it should perform, (2) an \textbf{input} that provides the detailed scenario description and the specific causal reasoning question, and (3) an \textbf{output} demonstrating a clear, structured, and comprehensive reasoning process explicitly based on the provided causal graph.


Below, we provide a detailed example from CausalDR dataset to illustrate precisely how we formulate these training samples. Each sample not only explicitly represents the causal graph, inference path, and answer but also cohesively integrates these components into a natural and intuitive explanatory narrative. This structured presentation is specifically designed to encourage LLMs to internalize a systematic approach to causal reasoning, thereby significantly enhancing inference consistency and reducing hallucinations.

\begin{tcolorbox}[breakable, colback=gray!5, colframe=gray!75!black, title=Fine-tuning Dataset Example (JSON), fonttitle=\bfseries]
\ttfamily\footnotesize
\begin{flushleft}
\{ \\
\ \ \ \textcolor{blue}{"instruction"}: "Given the question below, please construct a causal graph to analyze the scenario. Then, based on the causal graph, provide a detailed explanation of the step-by-step causal reasoning process. Finally, give the answer to the question based on the causal graph and the reasoning process.",\\

\ \ \ \textcolor{blue}{"input"}: "Imagine a self-contained, hypothetical world with only the following conditions, and without any unmentioned factors or causal relationships: Husband has a direct effect on wife and alarm clock. Wife has a direct effect on alarm clock. For husbands that don't set the alarm and wives that don't set the alarm, the probability of ringing alarm is 8\%. For husbands that don't set the alarm and wives that set the alarm, the probability of ringing alarm is 54\%. For husbands that set the alarm and wives that don't set the alarm, the probability of ringing alarm is 41\%. For husbands that set the alarm and wives that set the alarm, the probability of ringing alarm is 86\%. For husbands that don't set the alarm, the probability of alarm set by wife is 74\%. For husbands that set the alarm, the probability of alarm set by wife is 24\%. If we disregard the mediation effect through wife, would husband positively affect alarm clock?",\\

\ \ \ \textcolor{blue}{"output"}: "Alright, let me first review your input. Next, I will build a causal graph from the information provided, defining each node and clarifying how they interact. After that, I will detail the steps of causal inference, describing how I move from the causal graph to the final answer. To ensure clarity, I will begin by presenting the causal graph’s structure, the meaning of each node, and their connections. Then, I will illustrate the inference process, leading up to the result.\\

\textbf{\textit{Causal Graph:}}\\

First, here is the section on the causal graph nodes. For each node, I will list its ID, Name, and provide a brief description.\\

\textbf{Nodes}:\\
Node ID: X\\
Node Name: Husband\\
Node Description: Binary variable representing whether the husband sets the alarm (1=sets, 0=doesn't set)\\

Node ID: Y\\
Node Name: Alarm Clock\\
Node Description: Binary outcome variable representing whether the alarm rings (1=rings, 0=doesn't ring)\\

Node ID: V2\\
Node Name: Wife\\
Node Description: Binary variable representing whether the wife sets the alarm (1=sets, 0=doesn't set)\\

Next, I will explain how these nodes are linked in the causal graph, showing how information flows between them. I will go through each node in turn, indicating which nodes feed into it and which nodes it influences.\\

\textbf{Edges}:\\
Node: X
Inputs: N/A 
Outputs: ['V2', 'Y'] \\ 

Node: Y
Inputs: ['X', 'V2'] 
Outputs: N/A \\ 

Node: V2
Inputs: ['X'] 
Outputs: ['Y'] \\ 

Based on the current provided input information, \textbf{the inference goal} is that "\textit{Determine if Husband (X) has a positive direct effect on Alarm Clock (Y) when excluding mediation through Wife (V2)}"\\

Given this goal, the following describes the causal inference process.\\

\textbf{\textit{Causal Inference Process}}: The causal graph shows Husband (X) directly influences both Wife (V2) and Alarm (Y), while Wife (V2) also directly affects Alarm (Y). To isolate X's direct effect on Y, we calculate the natural direct effect by holding V2 constant at its X=0 distribution. Using the provided probabilities: When X=0, P(V2=1)=0.74 and P(V2=0)=0.26. The direct effect compares Y probabilities under X=1 vs X=0 for these fixed V2 values. For V2=0: 0.41 (X=1) - 0.08 (X=0) = +0.33. For V2=1: 0.86 (X=1) - 0.54 (X=0) = +0.32. Weighting these differences by V2's X=0 distribution gives (0.26*0.33) + (0.74*0.32) = 0.32. This positive result confirms Husband has a direct positive effect on Alarm Clock when excluding the Wife-mediated pathway.\\

\textit{\textbf{As a result of this causal inference process, I will reply with the answer yes}}."\\
\}
\end{flushleft}
\end{tcolorbox}

\subsection{Implementation Details of DeepSeek-R1}
\label{app:deepseek-implementation-details}

To construct the CausalDR dataset, we selected the reasoning large language model \textbf{DeepSeek-R1}~\cite{deepseekai2025} due to its strong reasoning capabilities and low inference cost. The model was deployed and accessed via the \textbf{DeepInfra API}\footnote{\url{https://deepinfra.com}}, an inference platform designed to streamline model integration. Following recommendations from DeepSeek-R1~\cite{deepseekai2025}, we configured the temperature at \textbf{0.6} and set the maximum token length to \textbf{8192} during dataset generation.

\subsection{Prompt Template for Generating CausalDR Data}
\label{app:structured-prompt-example}

Below, we provide the detailed prompt template used to guide the DeepSeek-R1 model in generating detailed causal graphs and explicit natural language reasoning paths for each training sample. 

\begin{tcolorbox}[colback=gray!5!white, colframe=gray!75!black, title=Detailed Prompt Template for Structured Causal Inference Data Generation, fonttitle=\bfseries]
\small
You are an expert specializing in causal inference and graph theory. Your task is to analyze a reasoning problem, construct a structured causal graph, and generate a detailed causal inference process. Your output must be in JSON format.

You will receive:
\begin{itemize}
  \item \textbf{Context \& Question:} A single block (\textcolor{blue}{\texttt{<context\_question> ... </context\_question>}}) that contains:
    \begin{itemize}
        \item Scenario description
        \item Constraints or rules
        \item Any additional details
    \end{itemize}
  \item \textbf{Reasoning:} This field contains the formal causal structure and mathematical \textcolor{blue}{reasoning} needed to solve the problem. It includes:
    \begin{itemize}
        \item Variable assignments (e.g., V1 = kraz, X = pexu)
        \item Causal graph structure notation (e.g., V1$\to$X, X$\to$Y)
        \item Probability calculations and mathematical steps required for the solution
    \end{itemize}
\end{itemize}

Your task:
\begin{itemize}
    \item Extract causal nodes and relationships from the provided input to construct a Causal Graph.
    \item Causal Graph is a Directed Acyclic Graph (DAG) that represents causal influences between different variables.
    \item Generate a structured causal reasoning process explaining how the conclusion is derived. In the \texttt{Causal Reasoning} field:
    \begin{itemize}
        \item \texttt{goal:} A concise statement of the reasoning question.
        \item \texttt{explanation:} A paragraph that:
        \begin{itemize}
            \item Describes variables and causal edges
            \item Explains causal influence propagation
            \item Translates formal math into intuition
            \item Justifies the final conclusion using probabilities
        \end{itemize}
    \end{itemize}
\end{itemize}

\textbf{Return JSON Format:}
\begin{verbatim}
{
  "Nodes": [
    {
      "id": "[DescriptiveVariableID]",
      "name": "[Variable Name]",
      "description": "[Detailed description of the causal variable]"
    }
    // ... more nodes if needed
  ],
  "Edges": [
    {
      "node": "[Same DescriptiveVariableID as in Nodes]",
      "inputs": ["List of all incoming causal nodes"],
      "outputs": ["List of all outgoing causal nodes"]
    }
    // ... Ensure that all Nodes are represented here.
  ],
  "Causal Reasoning": {
    "goal": "[Overarching question or objective]",
    "explanation": "[Step-by-step reasoning process]"
  },
  "Answer": "[yes/no]"
}
\end{verbatim}


\textcolor{blue}{\textless context\_question\textgreater}[\textit{Insert the given scenario description, constraints, and the specific question here.}\textcolor{blue}{\textless /context\_question\textgreater}

\textcolor{blue}{\textless reasoning\textgreater}[\textit{Insert the symbolic causal graph structure, variable assignments, and mathematical reasoning steps provided by the Cladder dataset here.}]\textcolor{blue}{\textless/reasoning\textgreater}
\end{tcolorbox}

\subsection{Detailed Algorithm for Fine-tuning Dataset Construction}
\label{app:algorithm-fine-tuning-data}

We provide a detailed pseudocode representation of our automated dataset construction pipeline in Algorithm~\ref{alg:fine_tuning_data_construction}. This algorithm clearly illustrates the structured process of generating high-quality fine-tuning data leveraging the DeepSeek-R1 model. Each step explicitly ensures data correctness, coherence, and suitability for supervised fine-tuning within our proposed framework.

\begin{algorithm}[htbp]
\caption{CausalDR Dataset Construction}
\label{alg:fine_tuning_data_construction}
\begin{algorithmic}[1]

\State \textbf{Input:} Cladder training set $D_{\text{Cladder}} = \{(c_i, q_i, r_i, a_i)\}_{i=1}^{N}$, where:
\Statex \quad $c_i$: scenario context;
\Statex \quad $q_i$: causal inference question;
\Statex \quad $r_i$: symbolic reasoning from Cladder dataset;
\Statex \quad $a_i$: ground-truth answer.
\Statex LLM for causal reasoning: DeepSeek-R1; maximum attempts $K=15$

\State \textbf{Output:} Fine-tuning dataset $D_{\text{CausalDR}}$

\State Initialize $D_{\text{CausalDR}} \gets \emptyset$

\ForAll{$(c_i, q_i, r_i, a_i) \in D_{\text{Cladder}}$}
    \State Construct structured prompt $p_i$ from $(c_i, q_i, r_i)$ (details in Appx.~A.4)
    \State Set $\text{success} \gets \text{False}$, $k \gets 0$
    \While{$\neg \text{success} \land k < K$}
        \State $(G_i, P_i, y_i) \gets \text{DeepSeek-R1}(p_i)$ \Comment{structured JSON output}
        \If{$y_i = a_i$}
            \State Construct coherent inference paragraph $S_i$ by explicitly integrating $(G_i, P_i, y_i)$ into a natural explanatory narrative (See example output in Appx.~A.2.). 
            \State $D_{\text{CausalDR}} \gets D_{\text{CausalDR}} \cup \{(c_i, q_i, S_i)\}$  \Comment{\(S_i\) encapsulates \(G_i, P_i, y_i\)}
            \State $\text{success} \gets \text{True}$
        \EndIf
        \State $k \gets k + 1$
    \EndWhile
\EndFor
\State \Return $D_{\text{CausalDR}}$

\end{algorithmic}
\end{algorithm}

Algorithm~\ref{alg:fine_tuning_data_construction} systematically describes how we integrate DeepSeek-R1 to produce a coherent reasoning sequence ($S$) comprising structured causal graphs ($G$), explicit inference paths ($P$), and answers ($y$), thereby ensuring the resulting fine-tuning dataset addresses the challenge of hallucinations in LLM inference.

\section{Additional Experimental Details}
\label{app:additional-experimental-details}

\subsection{Baselines Methods}
\label{appendix:reasoning_methods}
To assess the effectiveness of our \our{} method, we compare it against 5 commonly used reasoning methods. While these methods have achieved some success in reasoning, they still have limitations in dealing with complex causal reasoning tasks, including the difficulty of effectively capturing causal relationships, as described in the Introduction section. Specifically, our comparative approach consists of:
\begin{itemize}
    \item \textbf{Chain-of-Thought (CoT)}~\cite{wei2022chain}: CoT instructs the model to generate intermediate reasoning steps, helping models to solve complex problems by decomposing tasks into simpler sub-steps. We use zero-shot CoT without any few-shot prompting~\cite{kojima2023largelanguagemodelszeroshot}, only the reasoning prompt, and refer to it as CoT in the following sections.
    \item \textbf{Chain-of-Thought Self-Consistency (CoT-SC)}~\cite{wang2023self}: An improved version of CoT that samples multiple reasoning paths and selects the final answer based on consistency among these paths.
    \item \textbf{Causal Chain-of-Thought (CausalCoT)}~\cite{jin2023cladder}: CausalCoT guides the model through defined steps, including causal graph extraction, formalization of queries, and calculation of counterfactual outcomes.
    \item \textbf{Tree-of-Thoughts (ToT)}~\cite{yao2023tree}: ToT organizes the reasoning process as a tree structure, enabling the model to explore multiple reasoning paths. 
    \item \textbf{Graph-of-Thoughts (GoT)}~\cite{besta2024graph}: GoT  organizes reasoning steps into a graph structure, modeling each reasoning step as a node and dependencies among these steps as edges, without explicitly modeling causal relationships.
\end{itemize}

\subsection{Detailed Validation of Cladder Test Set}
\label{appendix:cladder_test_validation}

To ensure the quality and validity of our Cladder test set, we implemented a rigorous, two-step validation process involving both automated evaluation with DeepSeek-R1 and manual verification. Here, we describe this validation workflow in detail.

\paragraph{Step 1: Automated Validation via DeepSeek-R1.}  
We constructed prompts that asked the DeepSeek-R1 model to verify the correctness of the provided reasoning and answer for each test sample. An example validation prompt is as follows:

\begin{tcolorbox}[colback=gray!5!white, colframe=gray!75!black, title=Example Validation Prompt, breakable]
\small
\texttt{You are an expert analyzing causal reasoning. Evaluate if the reasoning process and answer are correct for this causal inference problem.}

\texttt{<context\_question>}...[Scenario and causal question here]...\texttt{</context\_question>}

\texttt{<reasoning>}...[Provided symbolic reasoning steps]...\texttt{</reasoning>}

\texttt{<proposed\_answer>}...[Provided answer]...\texttt{</proposed\_answer>}

Provide a JSON response:
\begin{verbatim}
{
  "reasoning_valid": true/false,
  "reasoning_error": "Brief description of error if any, otherwise 'None'",
  "answer_correct": true/false,
  "correct_answer": "yes/no",
  "brief_explanation": "1-2 sentences explaining your assessment"
}
\end{verbatim}
\end{tcolorbox}

If either the reasoning or the answer was marked incorrect, these samples were flagged for further review.

\paragraph{Step 2: Manual Verification.}  
Samples flagged as problematic by DeepSeek-R1 underwent manual review by domain experts to confirm the validity of the model's assessment. During this review, we carefully inspected reasoning accuracy and answer correctness, retaining only those samples unanimously confirmed as valid and logically sound.

Through this rigorous validation pipeline, we removed a total of 189 problematic samples, refining our test set down to 1,906 high-quality examples, suitable for rigorous causal reasoning evaluation.

The final validated Cladder test set, along with the validation scripts, will be publicly released to ensure reproducibility of our experiments.

\subsection{WIQA Data Preprocessing and Question Reformulation}
\label{appendix:wiqa-rewriting}

We first excluded irrelevant \textit{no-effect} perturbations, as these modifications are unrelated to the original causal chain and do not accurately assess a model’s causal reasoning capability. For efficiency and representativeness, we sampled 106 questions per subtask (212 in total; 95\% confidence, ±7\% margin). Subsequently, the selected questions were systematically reformulated to reduce ambiguity. Many original WIQA questions contained ambiguous phrasing or grammatical errors, potentially affecting evaluation results. An example of such ambiguity is:
\begin{tcolorbox}[colback=gray!5!white, colframe=gray!75!black, boxrule=0.5pt, arc=4pt, boxsep=5pt]
\textbf{Original Question (Ambiguous Example):}

\textit{``Suppose less DNA available happens, how will it affect hurting the DNA to replicate properly?''}

Options: A) more, B) less, C) no effect
\end{tcolorbox}

To eliminate such issues, we reformulated all questions into a clear, standardized format, strictly matching the provided answer options. The improved version of the above question is:

\begin{tcolorbox}[colback=gray!5!white, colframe=gray!75!black, boxrule=0.5pt, arc=4pt, boxsep=5pt]
\textbf{Improved Question (Reformulated Example):}

\textit{``Will having less available DNA cause more replication errors, fewer replication errors, or have no effect?''}

Options: A) more, B) less, C) no effect
\end{tcolorbox}

We used an automated approach employing the DeepSeek-R1 to rewrite each selected WIQA question. Below is the prompt we employed for the automatic rewriting:

\begin{tcolorbox}[colback=gray!5!white, colframe=gray!75!black, title=Prompt Used for Question Reformulation, fonttitle=\bfseries]

I have an English multiple-choice question with incorrect grammar and unclear meaning. I know that the correct answer is "\{Answer choice\}". Please help me rewrite this question so that:

- It is grammatically correct. \\
- It is logically clear and specific.\\
- It introduces no new information from outside the paragraph. \\
- It strictly preserves the original multiple-choice options: \\
  A) more, B) less, C) no effect.\\
- The question must be rewritten exactly in the following format: \\
  "Will [cause/change] cause more [effect], fewer [effect], or have no effect?"

This format must match the options exactly and avoid ambiguity. The rewritten question should clearly express the potential impact of a change on a specific outcome, and the options "more", "less", and "no effect" should directly correspond to the parts of the question.\\

Here is the background context:\\

Process steps:  \\
\{List of paragraph steps provided here\}\\

Original question:  \\
\{Original problematic question provided here\}\\

Options:  \\
A) more  \\
B) less  \\
C) no effect\\

Correct answer: \{Correct answer choice provided here\}\\

Return your result strictly in the following JSON format:\\
\{\\
  "improved\_question": "Your improved question here"\\
\}
\end{tcolorbox}

All 212 selected WIQA questions (106 from INPARA\_EFFECT and 106 from EXOGENOUS\_EFFECT) underwent this reformulation. A random subset of reformulated questions was manually reviewed to confirm grammatical correctness and logical clarity.

\subsection{Details of LLM-based Evaluation for Causal Graph Quality}
\label{appendix:llm_eval}
We employed an automatic evaluation approach utilizing an LLM (\texttt{gpt-4o-mini}) as a judge to assess the quality of the generated causal graphs. Specifically, given a causal reasoning context, a ground-truth causal graph, and a model-generated causal graph, the evaluator rated each generated graph along three dimensions: \textbf{Node Accuracy}, \textbf{Edge Accuracy}, and \textbf{Structural Fidelity}, assigning scores on a scale from 0 to 10. To ensure reproducibility, we set the inference temperature of \texttt{gpt-4o-mini} to 0.

\subsubsection{Causal Graph Quality Scoring Criteria}

The detailed scoring criteria for these dimensions are presented in Table~\ref{tab:scoring_criteria}.

\begin{table}[!htbp]
    \centering
    \caption{Detailed scoring criteria for LLM-based causal graph quality evaluation.}
    \resizebox{0.9\columnwidth}{!}{
    \begin{tabular}{c|p{4cm}|p{4cm}|p{4cm}}
        \toprule
        \textbf{Score} & \textbf{Node Accuracy} & \textbf{Edge Accuracy} & \textbf{Structural Fidelity} \\
        \midrule
        10 & All nodes perfectly identified, no errors or omissions. & All edges (including directions) identified perfectly. & Structure perfectly matches the Ground Truth; fully reasonable.\\ \hline
        9 & All core nodes correctly identified; only minor discrepancies with non-critical nodes. & Nearly perfect; only one minor discrepancy on a non-critical edge. & Structure highly matches, minor irrelevant differences only.\\ \hline
        8 & Nearly all nodes correctly identified; only 1 minor node omitted or misidentified. & Nearly all edges correct; just 1 minor edge omitted or misidentified. & Structure largely matches; minor differences but no significant flaws.\\ \hline
        7 & Core nodes identified accurately, but minor omissions or misidentifications (1–2 non-critical nodes). & Most core edges correct; 1–2 non-critical edges missed or incorrect. & Clearly reasonable and coherent structure, minor noticeable flaws.\\ \hline
        6 & Most nodes correct, but clearly missing or misidentifying a few nodes. & Generally correct, but clearly missing or incorrectly identifying 1–2 important edges. & Generally reasonable but with clear structural errors or omissions.\\ \hline
        5 & Around half of the nodes correct; obvious omissions or errors. & Around half of the edges correct; obvious errors or omissions. & Obvious structural problems; overall logic still somewhat coherent.\\ \hline
        4 & Only a small portion of nodes correct; many omissions or errors. & Poorly identified; only a small portion of core edges correct. & Partially confusing; only some parts clearly reasonable. \\ \hline
        3 & Most nodes incorrect, only a few correct. & Mostly incorrect edges, only a few correct. & Mostly chaotic; few structurally reasonable elements.\\ \hline
        2 & Mostly incorrect; only one or two nodes correct by chance. & Only 1–2 edges correct. & Severe structural issues; only minor elements reasonable by chance.\\ \hline
        1 & Only one node identified correctly; all others wrong. & Almost entirely incorrect; only one edge correct by chance. & Nearly completely incorrect; minimal structural coherence by chance.\\ \hline
        0 & Completely incorrect; no correct nodes identified. & Completely incorrect; no correct edges identified. & Completely incorrect; no structural coherence at all.\\
        \bottomrule
    \end{tabular}}
    \label{tab:scoring_criteria}
\end{table}

\subsubsection{Prompt Template for LLM-based Evaluation}
The prompt template used for this automatic evaluation was as follows:
\begin{tcolorbox}[colback=gray!5!white, colframe=gray!75!black, title=Prompt Template for LLM-based Evaluation, fonttitle=\bfseries, breakable]
\small

You are an expert evaluator specialized in assessing the quality of causal graph structures.\\[4pt]

Your task:\\[2pt]
Given a specific causal reasoning scenario (the problem context), along with a Ground Truth causal graph description (serving as the evaluation standard), your goal is to evaluate the quality of a \textbf{model-generated causal graph} (the evaluation target).\\[4pt]

Important Clarifications:\\[2pt]
- You are to assign scores specifically to the model-generated causal graph, NOT to the Ground Truth causal graph.\\[2pt]
- Your evaluation must be strictly based on comparing the model-generated causal graph with the provided Ground Truth causal graph and guided by the causal reasoning problem context, which clarifies the meaning of each node and edge.\\[2pt]
- Evaluate separately along three independent dimensions:\\
\quad • Node Accuracy(0–10 points)\\
\quad • Edge Accuracy(0–10 points)\\
\quad • Overall Structural Quality(0–10 points)\\[2pt]
- Follow the detailed scoring criteria provided below, and briefly justify your rating for each dimension.

Detailed Scoring Criteria (0–10 points each dimension):\\[2pt]

Node Accuracy:\\
- 10: All nodes perfectly identified, no errors or omissions.\\
- 9: All core nodes correctly identified; only minor discrepancies with non-critical nodes.\\
- 8: Nearly all nodes correctly identified; only 1 minor node omitted or misidentified.\\
- 7: Core nodes identified accurately, but minor omissions or misidentifications (1–2 non-critical nodes).\\
- 6: Most nodes correct, but clearly missing or misidentifying a few nodes.\\
- 5: Around half of the nodes correct; obvious omissions or errors.\\
- 4: Only a small portion of nodes correct; many omissions or errors.\\
- 3: Most nodes incorrect, only a few correct.\\
- 2: Mostly incorrect; only one or two nodes correct by chance.\\
- 1: Only one node identified correctly; all others wrong.\\
- 0: Completely incorrect; no correct nodes identified.\\[4pt]

Edge Accuracy:\\
- 10: All edges (including directions) identified perfectly.\\
- 9: Nearly perfect; only one minor discrepancy on a non-critical edge.\\
- 8: Nearly all edges correct; just 1 minor edge omitted or misidentified.\\
- 7: Most core edges correct; 1–2 non-critical edges missed or incorrect.\\
- 6: Generally correct, but clearly missing or incorrectly identifying 1–2 important edges.\\
- 5: Around half of the edges correct; obvious errors or omissions.\\
- 4: Poorly identified; only a small portion of core edges correct.\\
- 3: Mostly incorrect edges, only a few correct.\\
- 2: Only 1–2 edges correct.\\
- 1: Almost entirely incorrect; only one edge correct by chance.\\
- 0: Completely incorrect; no correct edges identified.\\[4pt]

Overall Structural Quality:\\
- 10: Structure perfectly matches the Ground Truth; fully reasonable.\\
- 9: Structure highly matches, minor irrelevant differences only.\\
- 8: Structure largely matches; minor differences but no significant flaws.\\
- 7: Clearly reasonable and coherent structure, minor noticeable flaws.\\
- 6: Generally reasonable but with clear structural errors or omissions.\\
- 5: Obvious structural problems; overall logic still somewhat coherent.\\
- 4: Partially confusing; only some parts clearly reasonable.\\
- 3: Mostly chaotic; few structurally reasonable elements.\\
- 2: Severe structural issues; only minor elements reasonable by chance.\\
- 1: Nearly completely incorrect; minimal structural coherence by chance.\\
- 0: Completely incorrect; no structural coherence at all.\\[4pt]

Please strictly follow the JSON format below when returning your evaluation:\\[4pt]
\{\\
\quad "Node\_Accuracy": \{"Score": (0–10), "Brief\_Reasoning": "…"\},\\
\quad "Edge\_Accuracy": \{"Score": (0–10), "Brief\_Reasoning": "…"\},\\
\quad "Overall\_Structural\_Quality": \{"Score": (0–10), "Brief\_Reasoning": "…"\}\\
\}\\[4pt]

Now, proceed to your evaluation:\\[4pt]

Causal Reasoning Problem Context:\\
\{problem\_context\}\\[4pt]

Ground Truth Causal Graph Description (Evaluation Standard):\\
\{reasoning with ground truth causal graph\}\\[4pt]

Model-generated Causal Graph Description (Evaluation Target):\\
\{LLM's output for the problem (including causal diagram)\}\\

\end{tcolorbox}

\subsubsection{Example of LLM-based Evaluation of Causal Graph Quality}

To clearly illustrate the LLM-based evaluation procedure used in our experiments, we provide a detailed example. Below, we demonstrate step-by-step how we objectively assessed the quality of causal graphs generated by different training methods (BaseModel (Llama-3.1-8B), and \our{}-Enhanced).

\begin{tcolorbox}[colback=gray!5!white, colframe=gray!75!black, title=Detailed Evaluation Example, fonttitle=\bfseries, breakable]
\small
\textbf{(1) Causal Reasoning Problem Context:}

\begin{quote}
Imagine a self-contained, hypothetical world with only the following conditions, and without any unmentioned factors or causal relationships: Demand has a direct effect on supply and price. Yield per acre has a direct effect on supply. Supply has a direct effect on price. Demand is unobserved. The overall probability of increased supply is 60\%. The probability of reduced supply and increased price is 25\%. The probability of increased supply and increased price is 24\%. Is the chance of increased price smaller when observing increased supply?
\end{quote}

\vspace{0.5em}
\textbf{(2) Ground Truth Causal Graph (Evaluation Standard):}

\begin{itemize}[leftmargin=1.5em]
    \item \textbf{Nodes:} Let V2 = yield per acre; V1 = demand; X = supply; Y = price.
    \item \textbf{Edges:} V1 -\textgreater X, V2 -\textgreater X, V1 -\textgreater Y, X -\textgreater Y
\end{itemize}

\vspace{0.5em}
\textbf{(3) Model-generated Causal Graphs (Evaluation Targets):}

\underline{BaseModel (Llama-3.1-8B) Raw Output:}
\begin{quote}
To analyze the scenario, we need to construct a causal graph. 

The causal graph for the scenario is as follows:

A (Demand) -\textgreater B (Supply) \\ 
B (Supply) -\textgreater C (Price) \\ 
B (Supply) -\textgreater D (Yield per acre) 

From the given probabilities, we have the following conditional probability tables:
[omit reasoning parts...]
\end{quote}

\underline{\our{}-Enhanced (Llama-3.1-8B) Raw Output:}
\begin{quote}
Alright, let me first review your input. Next, I will build a causal graph from the information provided, defining each node and clarifying how they interact. After that, I will detail the steps of causal inference, describing how I move from the causal graph to the final answer. To ensure clarity, I will begin by presenting the causal graph’s structure, the meaning of each node, and their connections. Then, I will illustrate the inference process, leading up to the result.

Causal Graph:

First, here is the section on the causal graph nodes. For each node, I will list its ID, Name, and provide a brief description.

Nodes:
Node ID: V1
Node Name: Demand
Node Description: Unobserved variable representing demand. Directly influences both supply and price.

Node ID: Y
Node Name: Price
Node Description: Price level. Directly influenced by demand and supply.

Node ID: X
Node Name: Supply
Node Description: Supply level. Directly influenced by yield per acre and demand, and directly affects price.

Node ID: V2
Node Name: Yield per Acre
Node Description: Yield per acre. Directly influences supply.

Next, I will explain how these nodes are linked in the causal graph, showing how information flows between them. I will go through each node in turn, indicating which nodes feed into it and which nodes it influences.

Edges:
Node: V1
Inputs: N/A
Outputs: ['X', 'Y']

Node: Y
Inputs: ['V1', 'X']
Outputs: N/A

Node: X
Inputs: ['V1', 'V2']
Outputs: ['Y']

Node: V2
Inputs: N/A
Outputs: ['X']

[omit reasoning parts...]
\end{quote}
\vspace{0.5em}
\textbf{(4) LLM-based Evaluation Results:}

Below are the LLM-generated evaluation results, strictly adhering to our defined scoring criteria and JSON format described in Appendix~\ref{appendix:llm_eval}.

\underline{BaseModel Evaluation:}
\begin{lstlisting}
{
  "Node_Accuracy": {
    "Score": 6, 
    "Brief_Reasoning": "The model-generated graph correctly identifies the core nodes (Demand, Supply, Price) but misidentifies the relationship with Yield per acre, which should be a direct effect on Supply rather than an edge from Supply to Yield."
  },
  "Edge_Accuracy": {
    "Score": 5, 
    "Brief_Reasoning": "The model-generated graph has the correct edges for Demand to Supply and Supply to Price, but incorrectly includes an edge from Supply to Yield instead of Yield to Supply, leading to significant errors in edge identification."
  },
  "Overall_Structural_Quality": {
    "Score": 5, 
    "Brief_Reasoning": "The structure is somewhat coherent with the core relationships but contains critical misrepresentations, particularly regarding the directionality of edges and the inclusion of Yield per acre, which affects the overall logical flow."
  }
}
\end{lstlisting}


\underline{\our{}-Enhanced Evaluation:}
\begin{lstlisting}
{
  "Node_Accuracy": {
    "Score": 10, 
    "Brief_Reasoning": "All nodes (Demand, Price, Supply, Yield per Acre) are perfectly identified and accurately described."
  },
  "Edge_Accuracy": {
    "Score": 10, 
    "Brief_Reasoning": "All edges are correctly identified with the correct directions, reflecting the causal relationships as per the Ground Truth."
  },
  "Overall_Structural_Quality": {
    "Score": 10, 
    "Brief_Reasoning": "The structure of the model-generated causal graph perfectly matches the Ground Truth, with clear and coherent relationships among the nodes."
  }
}
\end{lstlisting}

\end{tcolorbox}

\noindent
The above example explicitly illustrates the exact procedure and transparency of our evaluation methodology, starting from raw model outputs, extracting structured causal graphs, and finally obtaining standardized LLM-based scores.

\subsection{Case Study Supplementary}

\begin{figure}[ht!]
\centering
\includegraphics[width=\linewidth]{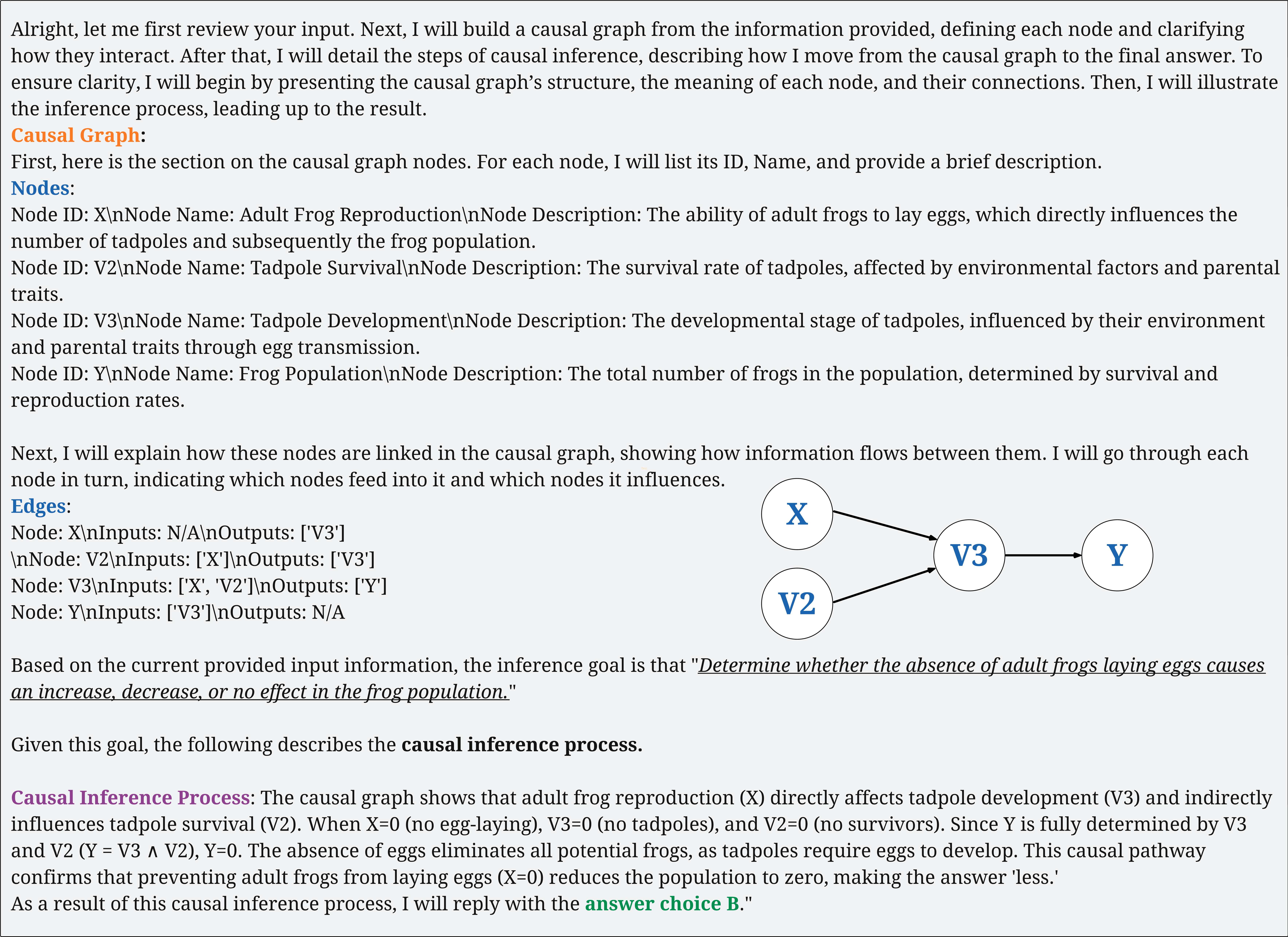}
\caption{\our{} inference trace demonstrating the output integrating causal DAG, reasoning path, and answer for a counterfactual reasoning query.}
\label{fig:case_study_full}
\end{figure}

To demonstrate the practical effectiveness of \our{}, we present a case illustration on the complete execution flow for a counterfactual reasoning query from WIQA that requires understanding complex causal mechanisms in Fig.~\ref{fig:case_study_full}.  The selected query exemplifies the challenges that \our{} addresses: "\textit{Process steps:  An adult frog spawns eggs in water $|$ These eggs hatch into tadpoles and continue to live in the water $|$ The tadpoles grow developing external gills and a longer tail $|$ The tadpole begins storing food in the tail $|$ The tadpole develops hind legs and lives off food stored in the its tail $|$ The front legs appear and the tadpoles tail shortens $|$ The tadpole now looks like a young frog $|$ The young frog continues to grow to adulthood. Question: "Will the absence of adult frogs laying eggs cause more frog population, less frog population, or have no effect?"  Choice A: more; Choice B: less; Choice C: no effect}", which demands exact node identification, edge direction, and graph-based inference to trace effects without adding false links. The causal DAG construction phase breaks down the scenario into nodes and edges. The reasoning phase builds on the DAG to trace paths step by step following the reasoning goal. Finally, we got the answer. 

